\documentclass[10pt,twocolumn,letterpaper]{article}

\usepackage{cvpr}              
\usepackage[accsupp]{axessibility} 

\usepackage[dvipsnames]{xcolor}
\usepackage{nicematrix,tikz}
\usepackage{algorithm}
\usepackage{algpseudocode}
\usepackage{graphicx}
\usepackage{bbm}
\usepackage[toc,title,page]{appendix}

\usepackage{color, colortbl}
\definecolor{Gray}{gray}{0.9}	
\newcolumntype{g}{>{\columncolor{Gray}}c}

\usepackage{colortbl}
\usepackage{multirow}
\usepackage{nicematrix,tikz}
\usepackage{booktabs}
\usepackage{subcaption} 
\NiceMatrixOptions
  {
    custom-line = 
     {
       letter = : ,
       command = dashedline , 
       ccommand = cdashedline ,
       tikz = dashed
     }
  }

\def\eg{e.g.,~}               

\newlength\paramarginsize
\newlength\figmarginsize
\newlength\secmarginsize
\newlength\figcapmarginsize
\newlength\tabcapmarginsize

\newcommand{\figref}[1]{Figure~\ref{fig:#1}}

\long\def\ignorethis#1{}

\usepackage{cite} 
\definecolor{cvprblue}{rgb}{0.21,0.49,0.74}
\usepackage[pagebackref,breaklinks,colorlinks,citecolor=cvprblue]{hyperref}

\def\paperID{14457} 
\def\confName{CVPR}
\def\confYear{2024}

\title{Open-Set Domain Adaptation for Semantic Segmentation}

\author{Seun-An Choe\thanks{Equal contribution}
\and 
Ah-Hyung Shin\footnotemark[1]
\and 
Keon-Hee Park
\and 
Jinwoo Choi\thanks{Corresponding authors}
\and 
Gyeong-Moon Park\footnotemark[2] \\ \\
Kyung Hee University, Yongin, Republic of Korea\\
{\tt\small \{dragoon0905, dkgud111, pgh2874, jinwoochoi, gmpark\}@khu.ac.kr}
}

\begin{document}
\maketitle
\begin{abstract}
Unsupervised domain adaptation (UDA) for semantic segmentation aims to transfer the pixel-wise knowledge from the labeled source domain to the unlabeled target domain. However, current UDA methods typically assume a shared label space between source and target, limiting their applicability in real-world scenarios where novel categories may emerge in the target domain. In this paper, we introduce Open-Set Domain Adaptation for Semantic Segmentation (OSDA-SS) for the first time, where the target domain includes unknown classes. We identify two major problems in the OSDA-SS scenario as follows: 1) the existing UDA methods struggle to predict the exact boundary of the unknown classes, and 2) they fail to accurately predict the shape of the unknown classes. To address these issues, we propose \textbf{B}oundary and \textbf{U}nknown \textbf{S}hape-Aware open-set domain adaptation, coined \textbf{BUS}. Our BUS can accurately discern the boundaries between known and unknown classes in a contrastive manner using a novel dilation-erosion-based contrastive loss. In addition, we propose OpenReMix, a new domain mixing augmentation method that guides our model to effectively learn domain and size-invariant features for improving the shape detection of the known and unknown classes. Through extensive experiments, we demonstrate that our proposed BUS effectively detects unknown classes in the challenging OSDA-SS scenario compared to the previous methods by a large margin.
The code is available at \url{https://github.com/KHU-AGI/BUS}.

\end{abstract}    
\vspace{-5mm}
\section{Introduction}
\label{sec:intro}

In semantic segmentation, a model predicts pixel-wise category labels given an input image. Semantic segmentation has a lot of applications, \eg autonomous driving~\cite{geiger2012we}, human-machine interaction~\cite{tsai2017deep}, and augmented reality. Over the past decade, there has been notable advancement in supervised semantic segmentation driven by deep neural networks~\cite{long2015fully,yu2015multi,chen2017deeplab,xie2021segformer}. However, supervised semantic segmentation requires pixel-level annotations, which are labor-intensive and costly to collect. To mitigate the challenges, unsupervised domain adaptation (UDA) has emerged. Many studies~\cite{tsai2018learning, vu2019advent,melas2021pixmatch,hoyer2022daformer,hoyer2022hrda,hoyer2023mic} leverage the already-labeled source data to achieve high performance on the unlabeled target data. Notably, synthetic datasets such as GTA5~\cite{richter2016playing} and SYNTHIA~\cite{ros2016synthia} which are automatically generated by game engines present valuable resources for UDA research.


\begin{figure}
\centering
  \begin{subfigure}[b]{0.48\columnwidth}
    \includegraphics[width=\textwidth]{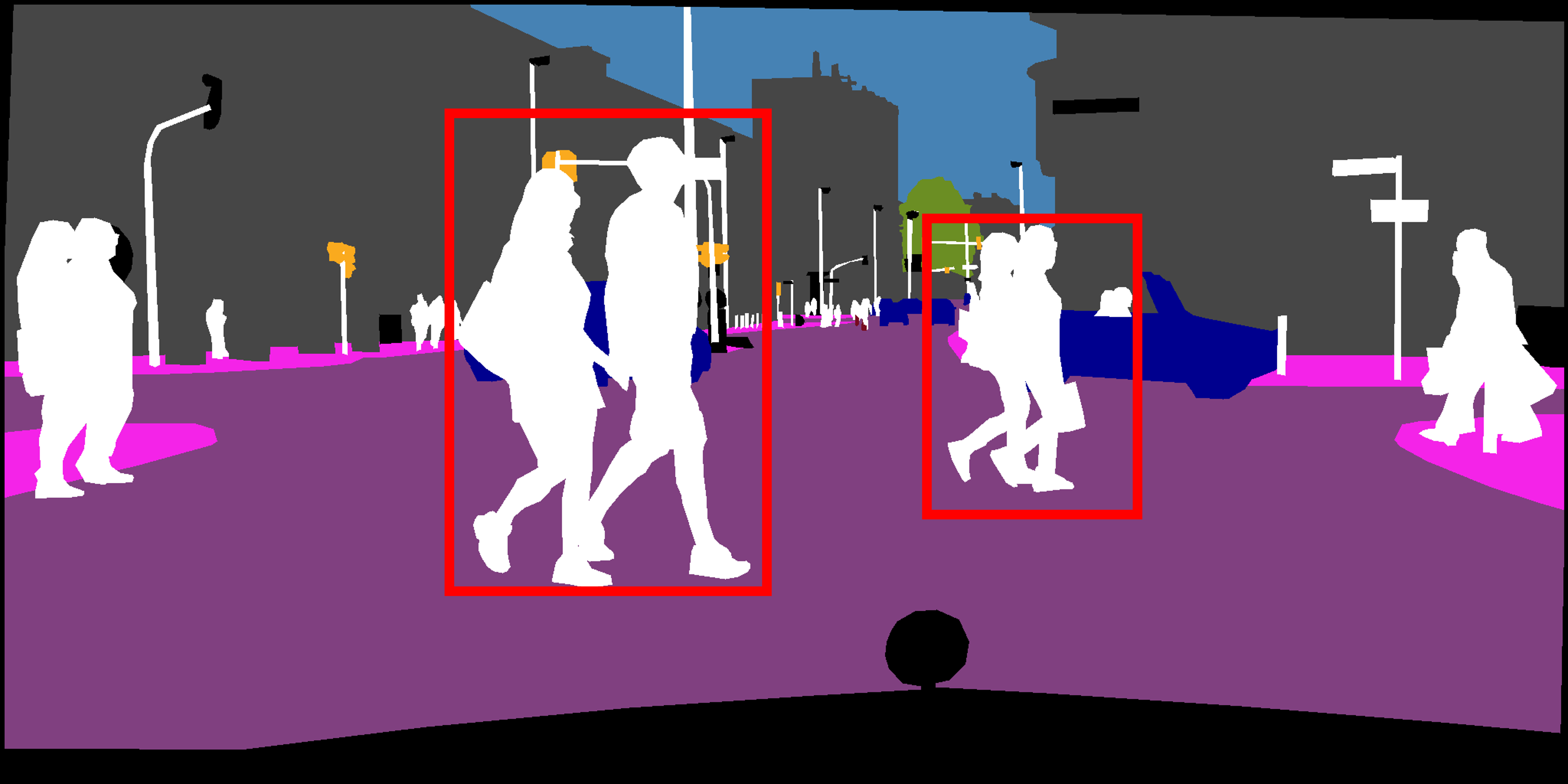}
    \caption{Ground truth.}
    \label{fig:1}
  \end{subfigure}
 \vfill
  \begin{subfigure}[b]{0.48\columnwidth}
    \includegraphics[width=\textwidth]{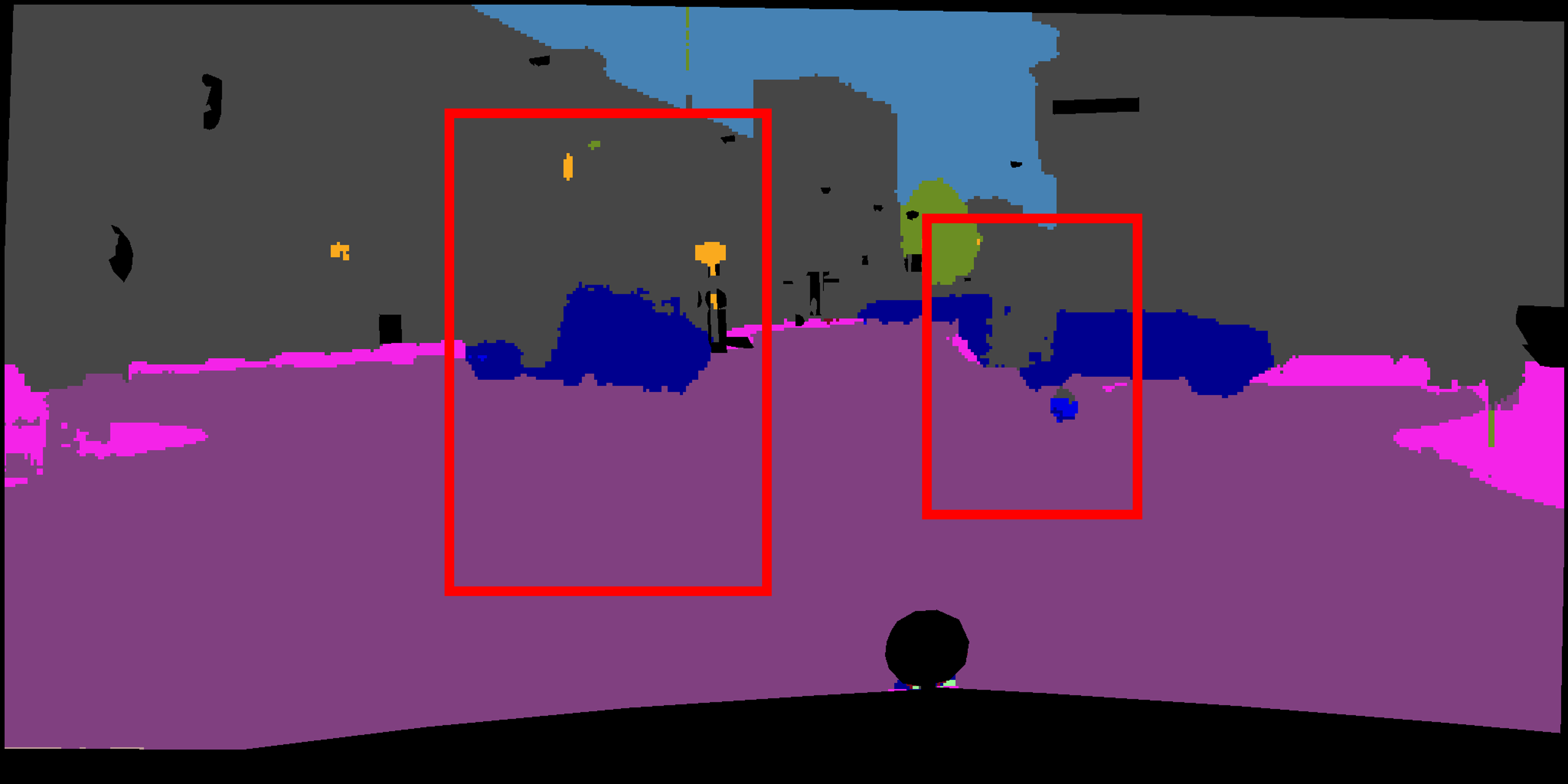}
    \caption{UDA method (MIC~\cite{hoyer2023mic}).}
    \label{fig:2}
  \end{subfigure}
  \hfill
  \begin{subfigure}[b]{0.48\columnwidth}
    \includegraphics[width=\textwidth]{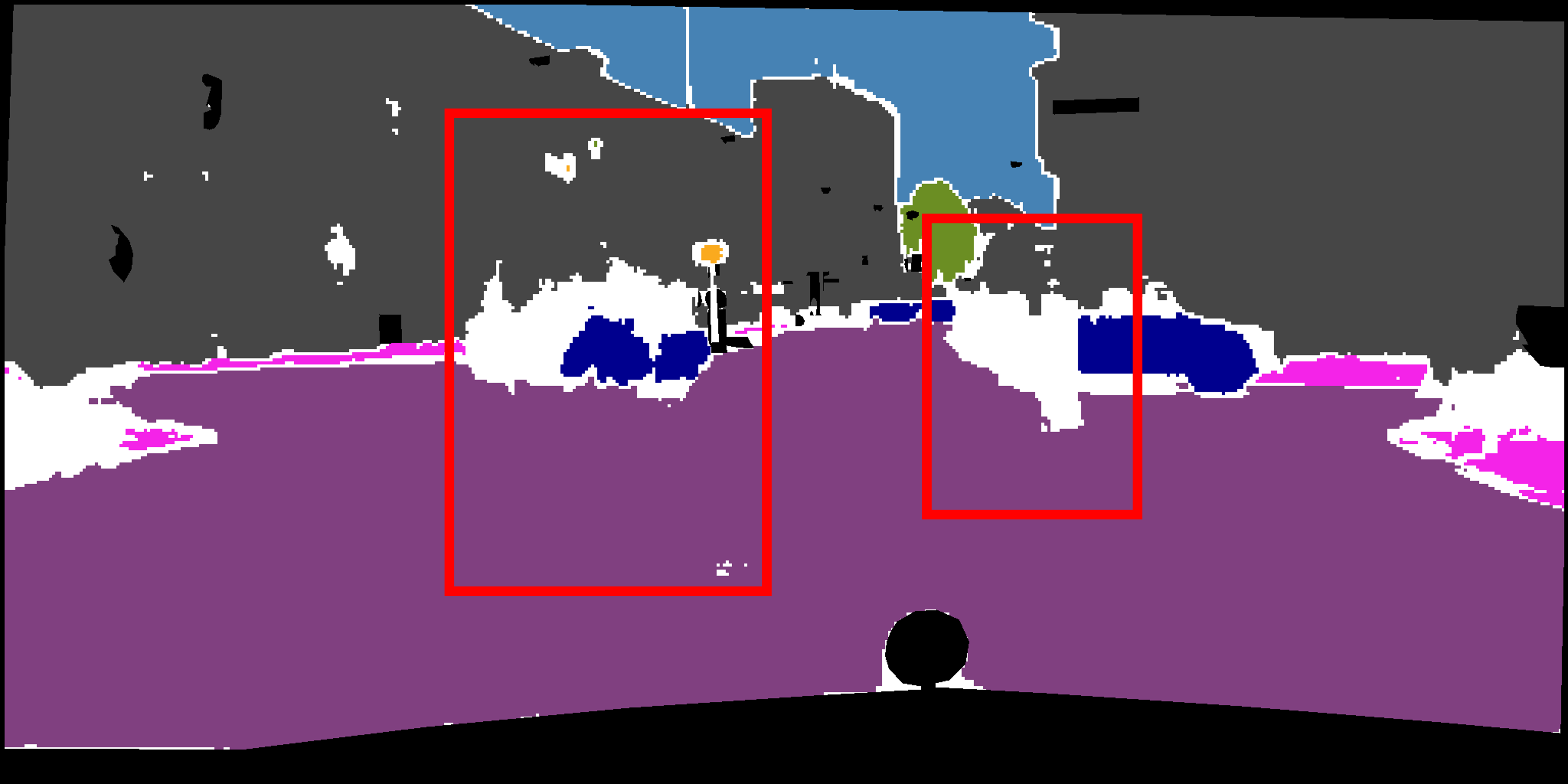}
    \caption{Confidence-based threshold.}
    \label{fig:3}
  \end{subfigure}
 \vfill
  \begin{subfigure}[b]{0.48\columnwidth}
    \includegraphics[width=\textwidth]{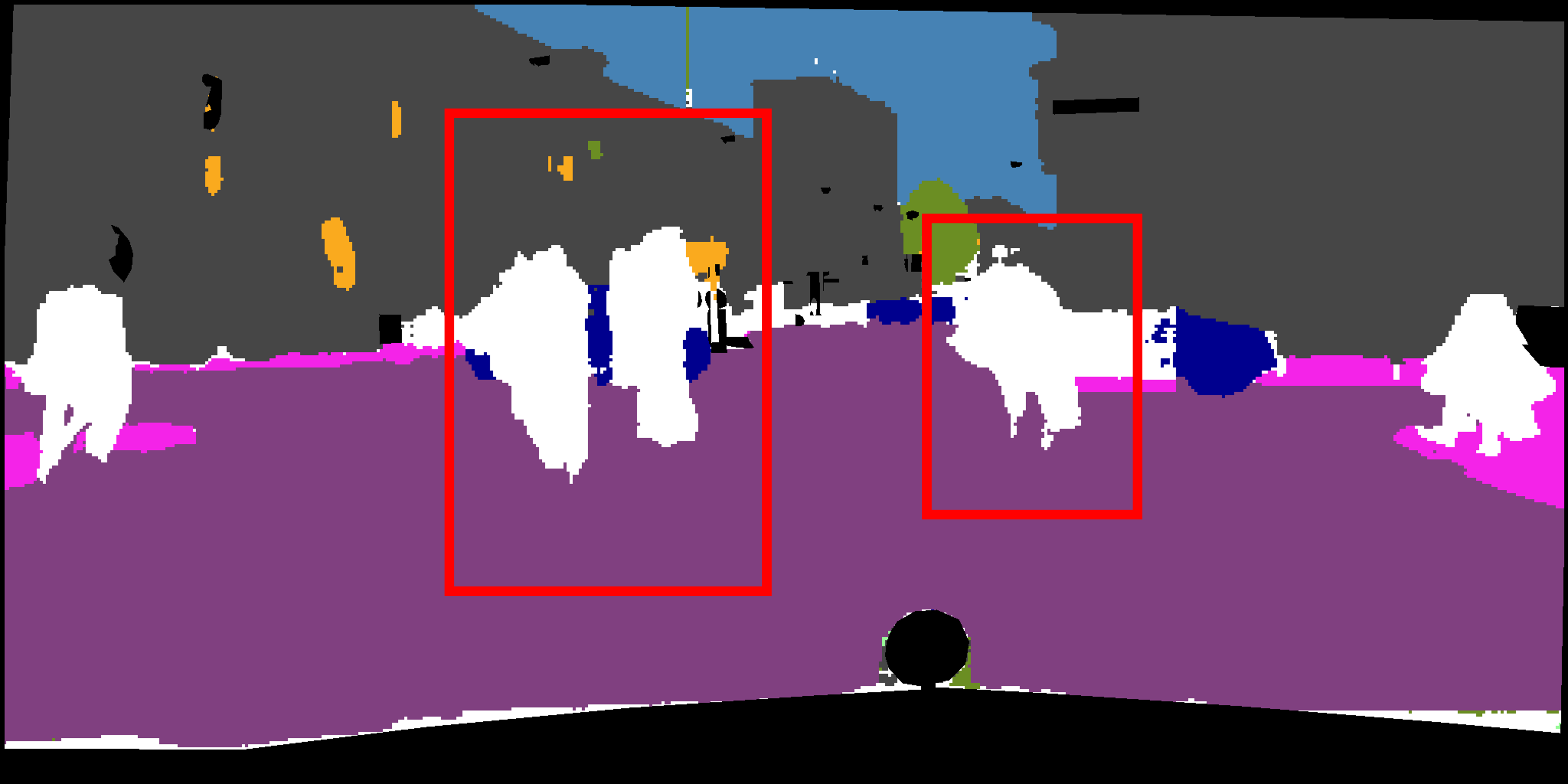}
    \caption{\small{Unknown head-expansion.}}
    \label{fig:4}
  \end{subfigure}
  \hfill
  \begin{subfigure}[b]{0.48\columnwidth}
    \includegraphics[width=\textwidth]{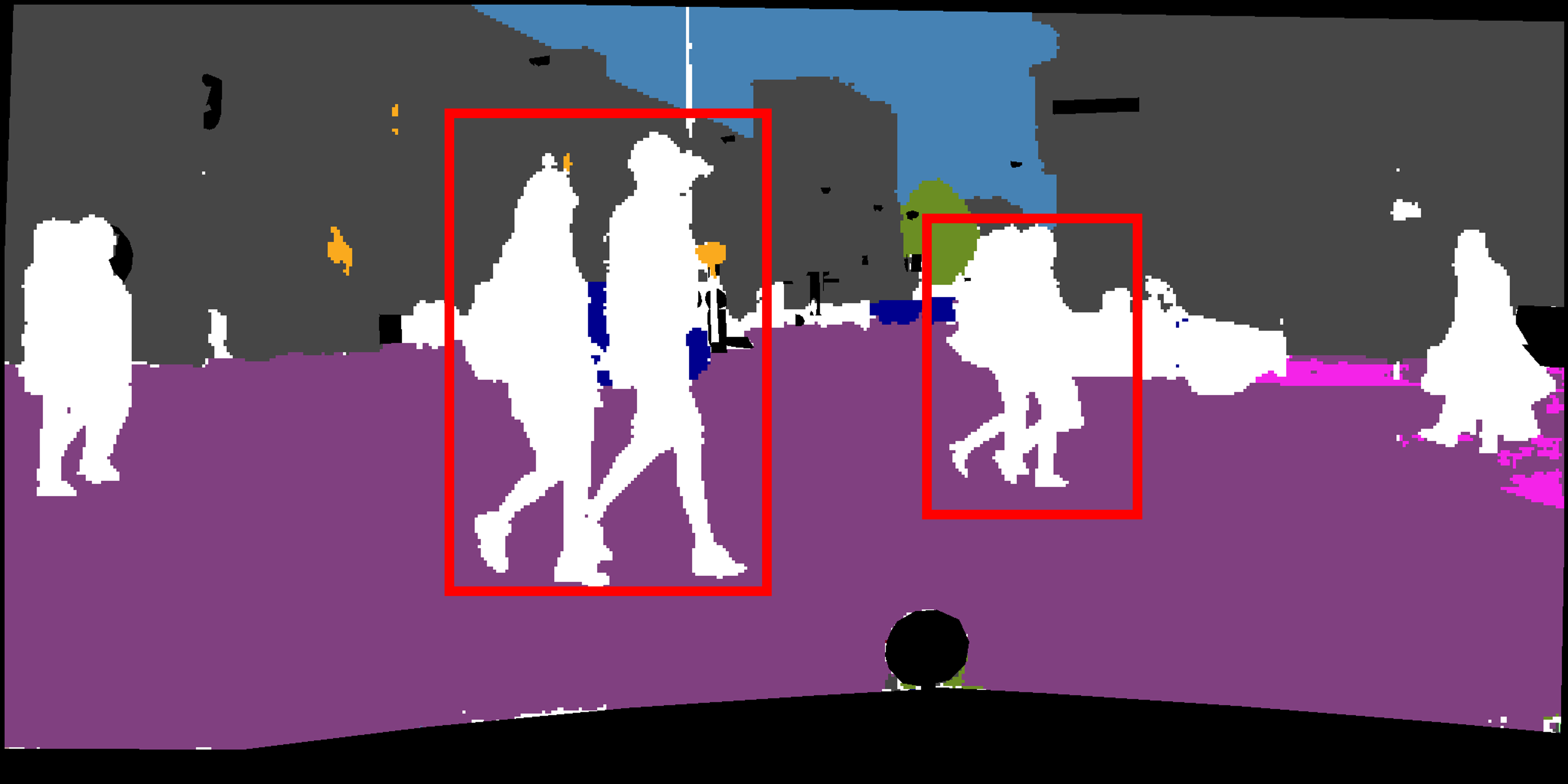}
    \caption{BUS (\textbf{Ours}).}
    \label{fig:5}
  \end{subfigure}
\vspace{-3mm}
\caption{Visualization of prediction maps in the OSDA-SS scenario. The pixels detected by the white color mean the unknown classes. The naive UDA method (b) is completely unaware of the \emph{unknown} classes. Even after applying simple techniques to help the UDA model recognize the \emph{unknown}, it still struggles to accurately predict the shape of the \emph{unknown}, as shown in (c) and (d).
\vspace{-6mm}}
\label{fig:motiv}
\end{figure}

UDA methods typically presume that source and target domains share the same label space. Such an assumption is not reasonable in real-world applications. In the target data, novel categories not presented in the source dataset (target-private categories) may emerge, leading to an Open-Set Domain Adaptation (OSDA) setting. The conventional UDA method may significantly fail under the OSDA setting, \eg a model erroneously label a person walking on the road as the road itself as shown in~\figref{motiv}(b). The desired model should reject any target-private classes as \emph{unknown} rather than misclassifying it as a known class. While OSDA has been widely explored in image classification~\cite{saito2018open, bucci2020effectiveness, jang2022unknown, you2019universal}, its application to semantic segmentation remains unexplored to the best of our knowledge. In this work, we tackle the interesting and challenging problem of Open-Set Domain Adaptation for Semantic Segmentation (OSDA-SS). Here, we deal with the labeled source data and the unlabeled target data containing classes not found in the source domain. In the OSDA-SS setting, the goal is to accurately predict pixel-wise category labels in the target domain and correctly distinguish the classes not seen during training as  \emph{unknown}.

One can design reasonable baselines by extending well-established UDA methods. One approach could be a confidence-threshold baseline. We train a model by using the UDA algorithm without considering target-private classes. During inference, 
the model identifies pixels with confidence scores below a predefined threshold as \emph{unknown}. We show the predicted segmentation map from the confidence-threshold baseline in~\figref{motiv}(c). Another baseline could be a head-expansion baseline. We expand the classification head from $C$ to $(C+1)$ dimensions, where $C$ represents the number of known classes. During training, when generating pseudo labels, we assign pixels with confidence scores lower than a specific threshold to the $(C+1)$-th head and train with the pseudo labels.
We show the predicted segmentation map from the head-expansion baseline in~\figref{motiv}(d). These baselines sometimes reject target-private classes as \emph{unknown}, but they often fail to do so, resulting in poor performance on the target dataset.

In this work, we build a model upon the head-expansion baseline. We find two failure modes of the baseline and propose a novel \textbf{B}oundary and \textbf{U}nknown \textbf{S}hape-Aware (\textbf{BUS}) OSDA-SS method. 
First, the previous models are often less confident or even fail near the boundaries of objects~\cite{liu2021bapa,caliva2019distance,li2020improving}. We find that the problem is even more severe for target-private classes due to lack of supervision. To address this issue, we propose a new \textbf{D}ilation-\textbf{E}rosion-based \textbf{CON}trastive (\textbf{DECON}) loss that manifests the boundaries through morphological operations, specifically dilation and erosion. 
Given a target image, we generate a target private mask using pseudo-labeling with the expanded head. 
Subsequently, we generate a boundary mask by subtracting the original private mask from the dilated private mask, indicating the region of \emph{known} classes near the boundaries.
We generate an erosion mask by applying erosion to the private mask, indicating more confident regions of the \emph{private classes}. We then train the model in a contrastive manner using the features from the erosion mask and the boundary mask as positive and negative samples, respectively. With DECON loss, our model clearly discerns the common and private classes near the boundaries.

Second, the baseline model faces challenges in accurately predicting the shape of \emph{unknown}. If the model consistently predicts the same object regardless of variations in size, it indicates that the model relies more on shape information than size information to recognize the object. Inspired by this motivation, we propose a new data mixing augmentation, \textbf{OpenReMix}. This method involves 1) resizing a random thing class from the source image and mixing it with the target image during training to consistently predict the same object even when its size varies. In addition, since there are no \emph{unknown} classes in the source image, 2) we cut the parts predicted as \emph{unknown} from a target image and paste them into a source image for supplemental learning of the last $(C+1)$-th head, aiding in the rejection of \emph{unknown} during source training. This delicate mixing strategy notably enhances the detection capability of \emph{unknown}, with a specific emphasis on capturing the shape information. By addressing the failure modes, the proposed BUS achieves significant performance gains on public benchmarks: GTA5 $\rightarrow$ Cityscapes and SYNTHIA $\rightarrow$ Cityscapes.

We summarize our major contributions as follows:
\begin{itemize}
\item To the best of our knowledge, we introduce a new task, Open-Set Domain Adaptation for Semantic Segmentation (OSDA-SS) for the first time. To tackle this challenging task, we propose a novel \textbf{B}oundary and \textbf{U}nknown \textbf{S}hape-Aware OSDA-SS method, coined \textbf{BUS}.

\item We introduce DECON loss, a new dilation-erosion-based contrastive loss to address the less confident and wrong predictions near the class boundaries.

\item We propose OpenReMix, which leads our model to learn size-invariant features and leverages \emph{unknown} objects from target to source to train the expanded head efficiently. OpenReMix encourages our model to focus on shape information of \emph{unknown} classes.

\item We conduct extensive experiments to validate the effectiveness of our proposed method. The proposed BUS shows state-of-the-art performance on public benchmark datasets with a significant margin.

\end{itemize}


\section{Related Work}
\label{sec:related}

\subsection{Semantic Segmentation.}


Semantic segmentation, which is a task to predict pixel-wise labels from the input images, has witnessed significant advances over the last decade. 
Key developments include fully convolution networks (FCNs)~\cite{long2015fully}, dilated convolution~\cite{chen2017deeplab,yu2015multi}, global pooling~\cite{liu2015parsenet}, pyramid pooling~\cite{zhao2018icnet,zhao2017pyramid,yang2018denseaspp}, and attention mechanism ~\cite{fu2019dual,huang2019ccnet,zhao2018psanet,zhu2019asymmetric}. 
Despite their success, these methods typically depend on a large amount of labeled data which is label-intensive and costly to collect.
In contrast, we formulate the semantic segmentation problem as domain adaptation to mitigate the annotation cost.

\subsection{Unsupervised Domain Adaptation for Semantic Segmentation.}
Recently, there has been a lot of work on unsupervised domain adaptation (UDA) for semantic segmentation.
UDA methods for semantic segmentation generally fall into two categories: adversarial learning-based and self-training approaches.
Adversarial learning-based methods~\cite{hong2018conditional,kim2020learning,pan2020unsupervised,tsai2018learning,tsai2019domain,chen2019synergistic,du2019ssf} utilize an adversarial domain classifier to learn domain-invariant representations, aiming to deceive the domain classifier.
Self-training methods~\cite{Zou_2018_ECCV,chen2019domain,zou2019confidence,wang2021domain,lian2019constructing,li2019bidirectional, Wang_2021_ICCV,melas2021pixmatch,zhang2021prototypical,tranheden2021dacs,hoyer2022daformer,hoyer2022hrda,hoyer2023mic} create pseudo labels for each pixel in the target domain image using confidence thresholding. 
Several self-training methods iteratively re-train the models, which result in enhanced performance on the target domain.
Despite the great success, most previous works assume a closed set setting, where the source and target domains share the same label space.
In this work, we relax this unrealistic assumption and tackle the problem of open-set domain adaptation for semantic segmentation (OSDA-SS). 
To the best of our knowledge, there is no prior work to tackle this problem.

\subsection{Open-Set Domain Adaptation}

Open-set domain adaptation (OSDA) extends UDA to handle novel categories in the target domain that are not present in the source domain. The primary goal of OSDA is to effectively distinguish the unknown categories from the known classes while reducing the domain gap between the source and target domains. Several OSDA methods have been proposed for the classification task~\cite{saito2018open, liu2019separate, wang2021progressively,jang2022unknown,mei2020instance}. However, in semantic segmentation task, which requires a higher degree of spatial information compared to classification, directly applying classification methods struggles to effectively differentiate unknown categories. The most similar work~\cite{bucher2021handling} to our method also deals with the novel classes that do not exist in the source domain. However, it accesses pre-defined private category definitions. To address this challenge, we propose a novel OSDA-SS task to discriminate unknown categories without needing to know any information about pre-defined class definitions.

\subsection{Domain Mixing Augmentation.}
To improve the generalization power of deep neural networks, mixup~\cite{zhang2017mixup,tokozume2018between} and its variants~\cite{olsson2021classmix,yun2019cutmix,french2020milking,tranheden2021dacs,gao2021dsp,berthelot2019mixmatch,french2019semi,na2021fixbi,wu2020dual,xu2020adversarial} have been proposed.
Especially, domain mixing augmentation demonstrates significant performance improvement in UDA~\cite{olsson2021classmix,yun2019cutmix,french2020milking,tranheden2021dacs,gao2021dsp, chen2022deliberated} by utilizing domain-mixed images as training data to encourage  learning of domain-invariant feature representations.
We propose OpenReMix, aiming to empower our model in capturing shape information, notably for the \emph{unknown} classes.
\nocite{moon2023online,seo2023lfs}


\begin{figure*}[t] 

\centering
\includegraphics[width=1\textwidth]{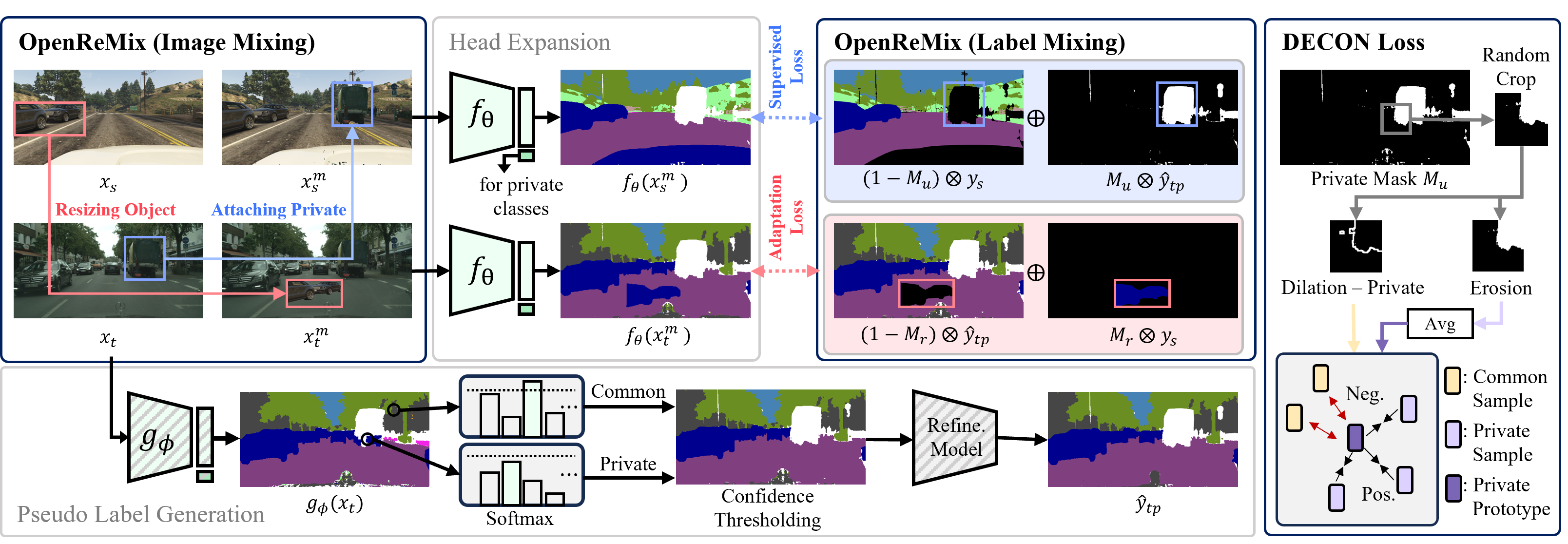}
\vspace{-6mm}

\caption{Overview of our proposed Boundary and Unknown Shape-Aware (BUS) method. 
We generate the mixed source image $x_{s}^m$ and the mixed target image $x_{t}^m$ from OpenReMix. The model is trained using the mixed source label and the mixed target pseudo-labels with supervised loss and adaptation loss, respectively. Especially, the expanded head is trained with the parts that predicted as unknown in pseudo-labels. Pseudo-labels are generated by thresholding the softmax probability and passing through the refinement network. DECON loss utilizes the dilation and erosion operations to distinguish the known and unknown classes near the boundaries.
\vspace{-3mm}}
\vspace{-3mm}
\label{fig:long}

\label{fig:onecol}

\end{figure*}


\section{Method}
\label{sec:Method}

\subsection{Problem Formulation}

In this section, we formulate a novel OSDA-SS task for the first time. In OSDA-SS, a network is trained with the source images \(X_s\)$=\{x_s^1, x_s^2, ..., x_s^{i_s}\}$ and the corresponding labels \(Y_s\)$=\{y_s^1, y_s^2, ..., y_s^{i_s}\}$ to ensure effective performance in the target domain \(X_t\)$=\{x_t^1, x_t^2, ..., x_t^{i_t}\}$ without labels.
\(x_s^{i_s} \in \mathbb{R}^{3 \times H \times W}\) and \(y_s^{i_s} \in \mathbb{R}^{C \times H \times W}\) are the \({i_s}\)-th source domain image and the pixel-wise label. 
\(H\) and \(W\) are the height and width of the image, respectively, and \(C\) denotes the number of categories in the source domain. In the target domain, we only have the image \(x_t^{i_t} \in \mathbb{R}^{3\times H \times W}\) without the corresponding labels. The source and target domains share \(C\) categories, and the target domain has additional unknown classes, \textit{i.e.}, the target images contain unknown objects. In this setting, the goal of OSDA-SS is to train a segmentation model \(f_\theta\) using both the labeled source data $(X_s, Y_s)$ and the unlabeled target data \(X_t\), and eventually the learned model \(f_\theta\) should predict both known and unknown classes well on the target domain.


\subsection{Baseline}
\label{sec:Baseline}
Inspired by the UDA methods based on self-training~\cite{tranheden2021dacs,hoyer2022daformer,hoyer2022hrda,hoyer2023mic}, we build a OSDA-SS baseline by extending the number of classifier heads from $C$ to $(C+1)$, where the $(C+1)$-th head corresponds to \emph{unknown} classes. The segmentation network \(f_\theta\) is trained with the labeled source data using the following categorical cross-entropy loss $\mathcal{L}_{seg}^{s}$:
\begin{align}
\mathcal{L}_{seg}^{s} = -\sum_{j=1}^{H \cdot W}\sum_{c=1}^{C+1} {y}_{s}^{(j,c)} \log f_\theta(x_{s})^{(j,c)},
\label{eq:source_cleanloss}
\end{align}
where $j \in \{1, 2, ..., H \cdot W\}$ denotes the pixel index and $c \in \{1, 2, ..., C+1\}$ denotes the class index. To alleviate the domain gap between the source and the target domains, the baseline utilizes a teacher network \(g_\phi\) to generate the target pseudo-labels. The pseudo-label $\hat{y}_{tp}^{(j)}$ for the $j$-th pixel considering \emph{unknown} is acquired as follows:
\begin{align}
\hat{y}_{tp}^{(j)} = 
\begin{cases} 
c', & \text{if } \left( \max_{c'}g_\phi(x_t)^{(j,c')} \geq \tau_p \right) \\
C+1, & \text{otherwise}
\end{cases}
,
\end{align}
where $c^{'} \in \{1, 2, ..., C\}$ denotes a class belonging to known classes and $\tau_p$ is a threshold. Using the above equation, we assign the less confident pixels as the \textit{unknown} class when the maximum softmax probability is lower than $\tau_p$.
Since we cannot completely trust the pseudo-labels above, we estimate the confidence of the pseudo-label by utilizing the ratio of confident pixels~\cite{tranheden2021dacs}. To this end, we count the number of pixels that have the maximum probability values exceeding a certain threshold \(\tau_t\) as follows:
\begin{align}
q_t = \frac{1}{H \cdot W} \sum_{j=1}^{H \cdot W} \left[ \max_{c'} g_\phi(x_t)^{(j, c')} \geq \tau_t \right],
\end{align}
where \(q_t\) means the confidence of the pseudo-label for the image. The network \(f_\theta\) is trained using the pseudo-labels and the corresponding confidence estimates with the following categorical cross-entropy loss $\mathcal{L}_{seg}^{t}$:
\begin{align}
\mathcal{L}_{seg}^{t} = -\sum_{j=1}^{H \cdot W}\sum_{c=1}^{C+1} {q_t \hat{y}_{tp}^{(j,c)} \log f_\theta(x_{t})^{(j,c)}}.
\label{eq:target_loss}
\end{align}
Finally, we update the teacher network \(g_\phi\) from \(f_\theta\) using the exponential moving average (EMA)~\cite{tarvainen2017mean} with a smoothing factor \(\alpha\) at the \((t+1)\)-th iteration, where the equation is shown as follows:
\begin{align}
\phi_{t+1} = \alpha \phi_t + (1 - \alpha) \theta_t.
\end{align}

Based on this baseline, we propose a novel \textbf{B}oundary and \textbf{U}nknown \textbf{S}hape-Aware OSDA method, coined \textbf{BUS}, which involves a new loss function to manifest the boundaries of known and unknown classes (see Section~\ref{sec:DECON}) and a new domain mixing augmentation to detect the shape of unknown objects robustly (see Section~\ref{sec:OpenReMix}).



\subsection{Dilation-Erosion-based Contrastive Loss}
\label{sec:DECON}
Semantic segmentation models often struggle to confidently predict object boundaries~\cite{liu2021bapa,caliva2019distance,li2020improving}, especially for target-private classes, where the absence of label information makes boundary prediction even more challenging. Since the models predict the boundaries with low confidence estimates, the quality of the generated pseudo-labels may not be accurate. If the model can confidently identify the boundaries of unknown classes, accurate predictions of unknown classes become feasible. 

To discern the boundaries effectively, we leverage two morphological operations, which are dilation and erosion. First, we utilize the pseudo-labels of the target image to create a target private mask as follows: 
\begin{align}
M_{u}^{(j)} = 
\begin{cases} 
1, & \text{if } \hat{y}_{tp}^{(j)}=C+1 \\
0, & \text{otherwise}
\end{cases},
\label{eq:target_private}
\end{align}
where $j$ denotes the pixel index.
Next, we apply the dilation function \(h_{d}(\cdot)\) and the erosion function \(h_{e}(\cdot)\) to the randomly cropped target private mask, generating dilation and erosion masks. In the dilation mask, we subtract the original target private mask to identify the regions associated with the common classes near the boundaries. On the other hand, the erosion mask emphasizes the regions that definitively belong to the private class. We generate these masks by the following equations:
\begin{align}
M_{N} &= h_{d}(M'_{u}) - M'_{u} , \\
M_{P} &= h_{e}(M'_{u}),
\end{align}
where $M'_{u}=r(M_{u})$ and $r(\cdot)$ is a function of random crop. \(M_{N}\) and \(M_{P}\) denote the masks representing the common and private parts, respectively.
To construct a contrastive loss, we generate anchor, positive, and negative samples using these masks as follows:
\begin{align}
z_i&= \mathrm{avg}(M_{P} \odot f_\theta(x_{t})), \\
z_j&=M_{P} \odot f_\theta(x_{t}), \\
z_k&= M_{N} \odot f_\theta(x_{t}),
\end{align}
where $z_i$ is an anchor, $\mathrm{avg}(\cdot)$ denotes the average pooling layer, and $z_j$ and $z_k$ represent positive and negative samples, respectively. We utilize $z_i$ as a prototype calculated by the average of positive samples. Finally, we define the contrastive loss~\cite{chen2020simple} using $z_i$, $z_j$, and $z_k$ as follows:
{\small
\begin{align}
\mathcal{L}_{DECON} =-\log \left[{\sum_{p=1}^{N_p} \exp(z_i \cdot z_j^p / \tau)}/{\sum_{n=1}^{N_n} \exp(z_i \cdot z_k^n / \tau)}\right],
\end{align}
}
where \(\tau\) is a temperature parameter. $N_n$ and $N_p$ denote the number of negative and positive pixels. To sum up, the proposed $\mathcal{L}_{DECON}$ allows our model to better distinguish between common and private classes near the boundaries.

\subsection{OpenReMix}
\label{sec:OpenReMix}
\paragraph{Resizing Object.}
We identify that the head-expansion baseline model fails to accurately predict the shape of the private classes. We hypothesize that if a model consistently predicts the same object regardless of size variations, the model can accurately predict the shape of the object as well. To this end, we extend the domain mixing method Classmix~\cite{olsson2021classmix}, which selects half of the classes from the source and appends them to the target image to learn domain-invariant features. 
On top of the Classmix, we introduce an additional step where we select one more thing class from the source image, resize it, and paste it to the random location of a target image with resizing object mask $M_r$. The mixed target image contains the same objects as the source image, but the sizes of the objects are different. Therefore, the model learns not only domain-invariant representations but also size-invariant representations from the mixed target images and the source images. 
This extension enhances the robustness of the model to size variations, contributing to the accurate prediction of the shape of unknown classes leading to superior open-set domain adaptation performance.


\paragraph{Attaching Private.}
As described in Section~\ref{sec:Baseline}, to address the target private classes, we expand the segmentation head. The expanded head is trained with the target pseudo-labels which contain the private labels. However, since there are no private classes in the source image, we cannot utilize the source data to update the additional head of the model. To overcome this inefficiency in training, we copy the parts of target private classes and paste them into a source image.
Given a target image, we create a target private mask \(M_{u}\) as Eq.~\eqref{eq:target_private}. With the target private mask, we copy the private regions in the target image to a source image, resulting in a private class-mixed source image. Similarly, by combining the labels of the source and the pseudo-labels of the target, we generate mixed source labels. This augmentation offers a significantly larger dataset for training to reject private classes, leading to improved open-set domain adaptation performance.
We formalize the attaching private process as follows. We generate a mixed source image $x_{s}^m$ and the corresponding source label \(y_{s}^m\) using the following equations:
\begin{align}
x_{s}^m = M_{u} \odot x_t + (1 - M_{u}) \odot x_s, \\
y_{s}^m = M_{u} \odot \hat{y}_{tp} + (1 - M_{u}) \odot y_s,
\end{align}
where $x_t$ and $\hat{y}_{tp}$ denote the target image and its pseudo-label. The mixed image \(x_{s}^m\) and the mixed label \(y_{s}^m\) are applied to Eq.~\eqref{eq:source_cleanloss}, instead of the source image $x_s$ and the corresponding label $y_s$.
  

  
  
  
  
  
  
  
  


\begin{table*}[t]
\centering
\resizebox{\textwidth}{!}{%
\setlength{\tabcolsep}{2.pt}
\renewcommand{\arraystretch}{1.2}
\begin{tabular}{lcccccccccccccggg}
\specialrule{1.5pt}{1pt}{0pt}
\multicolumn{1}{c|}{Method}                                   & Road  & S.walk & Build. & Wall  & Fence & Light & Veget. & Terrain & Sky   & Car   & Bus   & M.bike & \multicolumn{1}{c|}{Bike}  & Common & Private & H-Score \\ \hline
\multicolumn{17}{c}{\textbf{GTA5 $\rightarrow$ Cityscapes}}                                                                                                                                                                                                                                            \\ \hline
\multicolumn{1}{l|}{OSBP~\cite{saito2018open}}    & 4.92 & 3.93  & 42.8  & 2.55 & 6.04 & 14.29 & 68.58  & 26.50   & 44.21 & 41.78 & 0.94 & 7.20  & \multicolumn{1}{c|}{3.42} &  20.55  &  4.49   &  7.34   \\
\multicolumn{1}{l|}{UAN~\cite{you2019universal}}     & 65.97 & 23.41  & 76.41 & 37.26 & 18.50 & 20.13  & 80.57   & 30.37 & 82.47 & 77.35  & 27.80 & 16.62 & \multicolumn{1}{c|}{0.00}  &  38.00  &  3.59    &  6.56     \\
\multicolumn{1}{l|}{UniOT~\cite{jang2022unknown}} & 17.67  & 5.14   & 44.86   & \textbf{55.45}  & 2.31  & 52.61  & 40.01   & 3.37    & 79.43  & 52.87  & 52.31  & 7.18   & \multicolumn{1}{c|}{0.00}  &  20.20  &  5.36    &  7.49    \\ \hline
\multicolumn{1}{l|}{ASN~\cite{tsai2018learning}}          & 82.34 & 2.21   & 75.30   & 8.01  & 3.52  & 9.99  & 71.96  & 15.61   & 70.97 & 77.16 & 22.59 & 20.8   & \multicolumn{1}{c|}{0.06}  &  35.43  &  10.84   &  16.60   \\
\multicolumn{1}{l|}{Pixmatch~\cite{melas2021pixmatch}}            & 79.27 & 2.06   & 72.36  & 6.96  & 2.94  & 11.07 & 76.29  & 23.23   & 77.72 & 79.77 & 44.72 & 18.02  & \multicolumn{1}{c|}{0.01}  &  38.03  &  9.46    &  15.15   \\
\multicolumn{1}{l|}{DAF~\cite{hoyer2022daformer}}            & 94.26 & 48.69  & 83.47  & 38.67 & 32.83 & 41.71 & 87.79  & 39.15   & \textbf{93.59} & 85.29 & 47.04 & 28.36  & \multicolumn{1}{c|}{46.86} &  61.26  &  14.63   &  23.36   \\
\multicolumn{1}{l|}{HRDA~\cite{hoyer2022hrda}}                & \textbf{95.14} & 62.58  & 82.92  & 47.44 & 43.57 & 53.18 & 88.26  & 44.42   & 92.92 & 90.23 & 57.43 & 14.71  & \multicolumn{1}{c|}{56.83} &  63.82  &  12.13   &  20.39   \\
\multicolumn{1}{l|}{MIC~\cite{hoyer2023mic}}                 & 93.26 & 58.96 & 79.30  & 21.62 & 31.41 & 39.32 & 85.48  & 31.94   & 91.64 & 88.16 & 44.77 & 47.64  & \multicolumn{1}{c|}{42.77} &  58.17  &  11.87   &  19.71   \\ \hline
\multicolumn{1}{l|}{BUS \textbf{(Ours)}}                & 95.06 & \textbf{66.65}  & \textbf{90.53}  & 55.37 & \textbf{55.38} & \textbf{57.20} & \textbf{91.12}  & \textbf{49.69}   & 92.96 & \textbf{93.50} & \textbf{68.81} & \textbf{58.73}  & \multicolumn{1}{c|}{\textbf{67.04}} &  \textbf{72.47}  &  \textbf{55.42}   &  \textbf{62.81}   \\

\specialrule{1.5pt}{1pt}{1pt}
\end{tabular}%
}
\vspace{-4mm}
\end{table*}

\begin{table*}[t]
\resizebox{\textwidth}{!}{%
\setlength{\tabcolsep}{2.pt}
\renewcommand{\arraystretch}{1.}
\begin{tabular}{cccccccccccccggg}
\specialrule{1.5pt}{1pt}{1pt}
\multicolumn{1}{c|}{Method}                  & Road  & S.walk & Build. & Wall  & Fence & Light & Veget. & Sky   & Car   & Bus   & M.bike & \multicolumn{1}{c|}{Bike}  & Common                                            & Private                                          & H-Score                                          \\ \hline
\multicolumn{16}{c}{\textbf{SYNTHIA $\rightarrow$ Cityscapes}}                                                                                                                                                                                                                                                                  \\ \hline
\multicolumn{1}{l|}{OSBP~\cite{saito2018open}}       &    6.71   &    9.49    &   49.83     &   0.70    &  0.0     &   0.76    &   26.03     &   36.91    &   20.04    &   4.76    &   2.90     & \multicolumn{1}{c|}{8.70}      &                  13.20                                 &         4.90                                         &           7.14                                       \\
\multicolumn{1}{l|}{UAN~\cite{you2019universal}}  &   33.24    &   19.03    &    71.49    &    4.02    &   0.05    &   14.34    &    75.78   &    81.06    &   53.88    &   19.34    &   8.14    &    \multicolumn{1}{c|}{21.84}      & 31.30 & 4.53 & 7.91 \\
\multicolumn{1}{l|}{UniOT~\cite{jang2022unknown}}      &  0.00     &   16.79     &  18.52      &  1.05     &  6.49     &  16.8     &  14.52      &   57.4    &  6.48     &   2.59    &   3.73     & \multicolumn{1}{c|}{3.88}     & 12.35 & 5.49 & 7.06 \\ \hline
\multicolumn{1}{l|}{ASN~\cite{tsai2018learning}} &   72.70    &    41.29    &    73.59    &    7.38   &   0.08    &   1.17    &    71.35    &    82.22   &    67.35   &    23.30   &    0.94   & \multicolumn{1}{c|}{20.56}      &      38.49                                             &  4.62                                                &8.25
\\
\multicolumn{1}{l|}{Pixmatch~\cite{melas2021pixmatch}}   & 74.16      & 8.15       & 76.21       & 0.01      & 0.0      & 5.64      & 44.15       & 63.76      & 44.66      & 17.27      & 0.13       & \multicolumn{1}{c|}{0.38}      & 26.30                                                  & 6.87                                                 & 11.00                                                 \\
\multicolumn{1}{l|}{DAF~\cite{hoyer2022daformer}}  & 70.10 & 39.65  & 83.09  & 22.75 & 4.66  & 41.19 & 81.56  & 91.79 & 84.36 & 51.13 & 43.78  & \multicolumn{1}{c|}{46.20} & 51.49                                             & 9.07                                             & 15.57                                            \\
\multicolumn{1}{l|}{HRDA~\cite{hoyer2022hrda}}       & 85.62 & 41.74  & 83.29  & 36.35 & 0.86  & 35.17 & 83.98  & 90.90 & 84.74 & 50.42 & 46.78  & \multicolumn{1}{c|}{58.33} & 54.68                                             & 12.68                                            & 20.82                                            \\
\multicolumn{1}{l|}{MIC~\cite{hoyer2023mic}}        & \textbf{88.31} & \textbf{70.71}  & 85.00  & 26.23 & \textbf{6.60}  & 35.27 & 84.80  & 91.41 & 81.47 & 53.62 & 55.39  & \multicolumn{1}{c|}{58.20} & 57.46                                             & 10.02                                            & 17.23                                            \\ \hline
\multicolumn{1}{l|}{BUS \textbf{(Ours)}}       & 86.85 & 43.49 & \textbf{89.35}  & \textbf{46.12} & 4.39  & \textbf{54.29} & \textbf{87.90}  & \textbf{92.49} & \textbf{91.46} & \textbf{61.23} & \textbf{58.11}  & \multicolumn{1}{c|}{\textbf{59.81}} & \textbf{64.62}                                             & \textbf{33.37}                                            & \textbf{44.01}   
\\
\specialrule{1.5pt}{1pt}{1pt}
\end{tabular}%
}
\vspace{-2mm}
\caption{\label{tab:table-name1} Performance on two different benchmarks. Our proposed BUS achieved the state-of-the-art performance with remarkable improvement in H-Score $+39.45\%$ against DAFormer in GTA $\rightarrow$ Cityscapes and $+23.19\%$ against HRDA in SYNTHIA $\rightarrow$ Cityscapes.
\vspace{-3mm}}
\vspace{-3mm}
\end{table*}

\section{Experiments}
\label{sec:results}

\begin{figure*}[t] 

\centering

\includegraphics[width=1\textwidth]{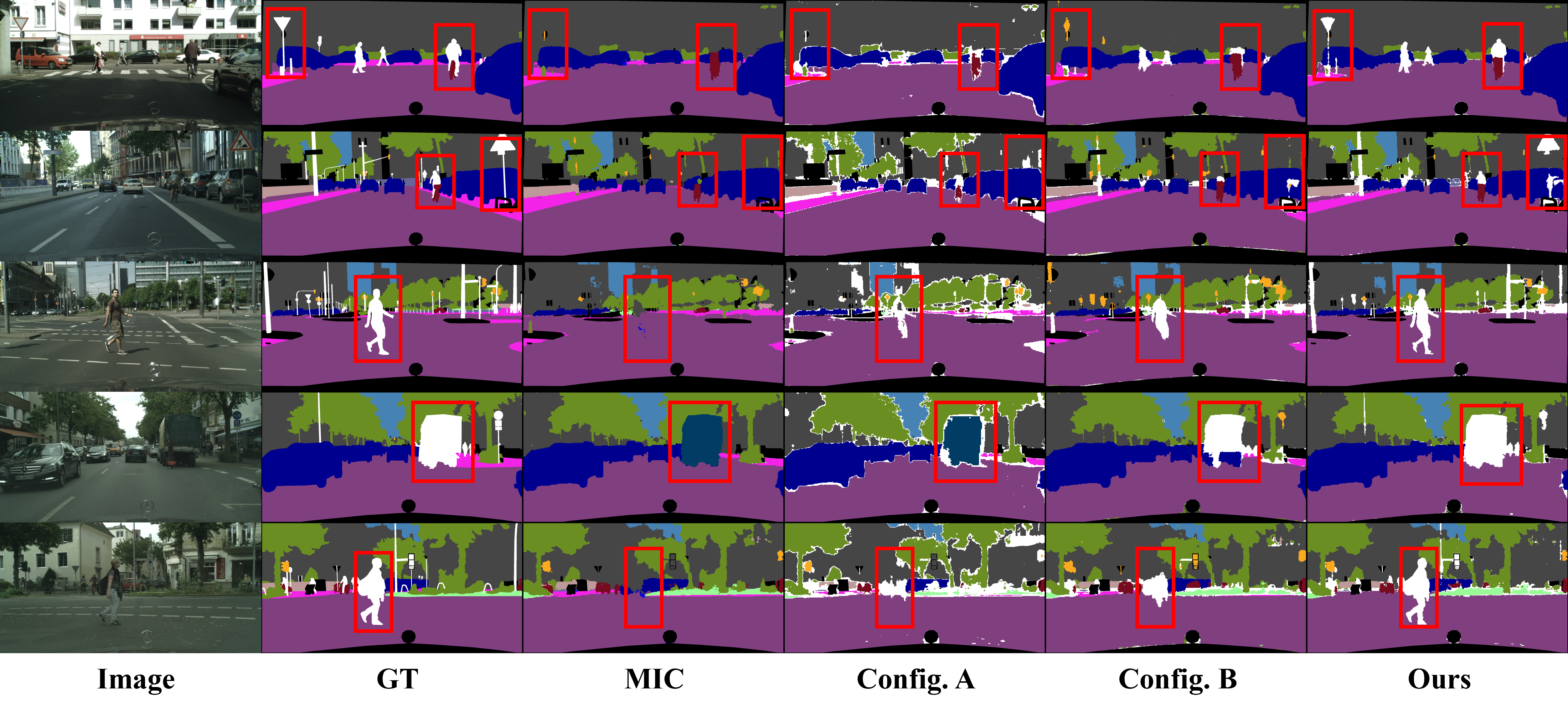}
\vspace{-8mm}
\caption{Qualitative comparison of our method with MIC, confidence-based MIC (Config.~A), and head-expansion (Config.~B) on the GTA5 $\rightarrow$ Cityscapes. GT represents the ground truth.
\vspace{-2mm}}

\label{fig:qu1}

\label{fig:onecol}

\end{figure*}

\begin{figure*}[t] 

\centering

\includegraphics[width=1\textwidth]{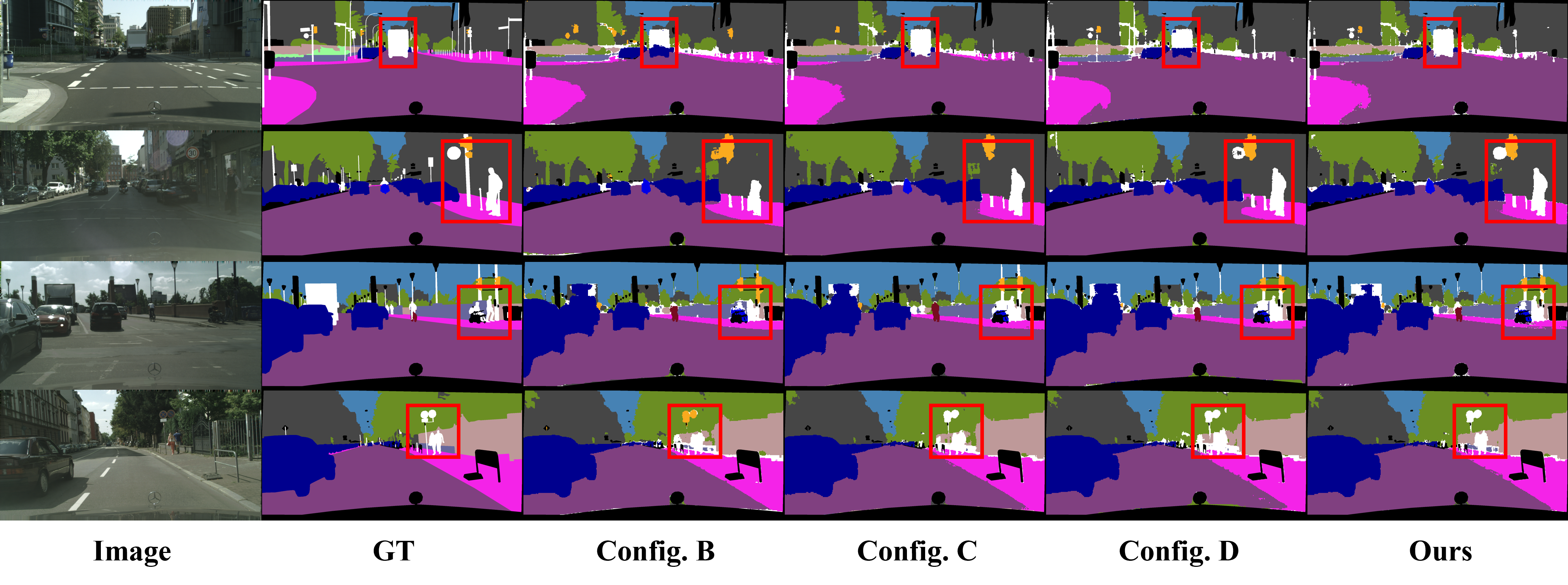}
\vspace{-8mm}
\caption{Qualitative comparison of our method with head-expansion (Config.~B), DECON loss (Config.~C), and OpenReMix (Config.~D) on the SYNTHIA $\rightarrow$ Cityscapes. GT represents the ground truth.
\vspace{-2.5mm}}
\vspace{-4mm}
\label{fig:qu2}

\label{fig:onecol}

\end{figure*}

\subsection{Experimental Setup}
\paragraph{Datasets.}
We evaluated our framework over two challenging synthetic-to-real scenarios in autonomous driving, i.e., GTA5 $\rightarrow$ Cityscapes and SYNTHIA $\rightarrow$ Cityscapes. GTA5~\cite{richter2016playing} is a synthesized dataset, which consists of 24,966 images with a resolution of $1914 \times 1052$. SYNTHIA~\cite{ros2016synthia} is also a synthesized dataset, which contains 9,400 images with resolution $1280 \times 760$. Cityscapes~\cite{cordts2016cityscapes} is a real-image dataset with 2,975 training samples and 500 validation samples with resolution $2048 \times 1024$. It shares 19 classes with GTA and 16 classes with SYNTHIA.
\paragraph{Scenario Construction.}
Using these datasets, we established new scenarios tailored for the OSDA-SS task. First, to create new classes emerging in the target domain that are not present in the source domain, we selected certain source classes to be removed. In autonomous driving scenarios, the classes that are likely to emerge in the target domain are expected to be ``thing" classes. Stuff classes representing the background area typically do not emerge as new classes. Therefore, we selected specific classes from the ``thing" categories to be excluded. The following list denotes the classes designated as unknown in GTA5 and SYNTHIA. 
\begin{itemize}
\item GTA5: ``pole", ``traffic sign", ``person", ``rider", ``truck", and ``train". \\
\vspace{-5mm}
\item SYNTHIA: ``pole", ``traffic sign", ``person", ``rider", ``truck", ``train", and ``terrain".
\end{itemize}
Notably, SYNTHIA includes the ``terrain", which inherently lacks labels from the outset.
Second, in order to avoid training the excluded classes, pixels corresponding to those classes were designated as ``ignore" and were not included in the loss function during training.
Finally, during the evaluation of the target domain, 
the above classes were treated as single unknown class.
\paragraph{Evaluation Metrics.}
Previous works~\cite{hoyer2022daformer,melas2021pixmatch,hoyer2022hrda,hoyer2023mic} used mIoU (mean Intersection-over-Union) as the evaluation metric, which averaged the IoU of each class. Since we treated every unknown classes as single unknown class, simply averaging would diminish the impact of private classes significantly. Therefore, inspired by~\cite{fu2020learning}, we utilized the harmonic mean of the mean IoU score for known classes (common) and the IoU score for one unknown class (private) as our evaluation metric, known as the H-Score.

\paragraph{Implementation Details.} 
We adopted DAFormer~\cite{hoyer2022daformer} network with the MiT-B5 encoder~\cite{xie2021segformer} pre-trained on imageNet-1K~\cite{deng2009imagenet}. We followed the multi-resolution self-training strategy and training parameters of MIC~\cite{hoyer2023mic}. The network was trained with AdamW~\cite{loshchilov2017decoupled}. The learning rates were set to 6e-5 for the backbone and 6e-4 for the decoder head, with a weight decay of 0.01 and linear learning rate warm-up over 1.5k steps. EMA factor was \(\alpha\)=0.999. We utilized the Rare Class Sampling~\cite{hoyer2022daformer}, ImageNet Feature Distance~\cite{hoyer2022daformer}, DACS~\cite{tranheden2021dacs} data augmentation, and Masked Image Consistency module~\cite{hoyer2023mic}. We trained on a batch of two $512 \times 512$ random crops for 40k iterations. We used MobileSAM~\cite{zhang2023faster} for the refinement model. The refinement process is described in the supplemental material.


\paragraph{Baselines.} 
We compared our approach with two scenarios. The first scenario comprised the Open-Set Domain Adaptation (OSDA) method like OSBP~\cite{saito2018open} and Universal Domain Adaptation (UniDA) methods like UAN~\cite{you2019universal} and UniOT~\cite{jang2022unknown}, which were capable of rejecting unknown classes but were primarily designed for classification tasks. 
The second scenario was Unsupervised Domain Adaptation (UDA) methods for semantic segmentation in closed-set setting, which included AdaptSegNet (ASN)~\cite{tsai2018learning}, Pixmatch~\cite{melas2021pixmatch}, DAFormer (DAF)~\cite{hoyer2022daformer}, HRDA~\cite{hoyer2022hrda}, and MIC~\cite{hoyer2023mic}. 
In the UDA method, we assigned the unknown label for regions with low confidence scores during inference. For OSDA and UniDA methods, we replaced the classification network with the DeepLabv2~\cite{chen2017deeplab} segmentation network, which uses ResNet-101~\cite{he2016deep} as the backbone, and adopted the image-level methods to the pixels.


\subsection{Comparison with the State-of-the-Art}
Table~\ref{tab:table-name1} showed the experimental results of GTA5 $\rightarrow$ Cityscapes and SYNTHIA$\rightarrow$ Cityscapes, respectively. The classification methods struggled to accurately discriminate the private classes in semantic segmentation tasks, which demanded a higher degree of spatial information. The UDA methods also faced challenges in effectively distinguishing private classes when simply leveraging a confidence-based approach.
In contrast, our proposed approach significantly outperformed the other comparison methods in H-Score. Especially, compared to the best baseline, our proposed BUS achieved a performance improvement of about $+39.45\%$ compared to DAF~\cite{hoyer2022daformer} in GTA $\rightarrow$ Cityscapes and about $+23.19\%$ compared to HRDA~\cite{hoyer2022hrda} in SYNTHIA $\rightarrow$ Cityscapes. This experiment demonstrated the effectiveness of our method in discriminating private classes while maintaining the performance of common classes.
A more detailed examination revealed that we achieved a significant improvement in the private class IoU score to approximately $+40.79\%$ compared to the DAF~\cite{hoyer2022daformer}, and also an increase in the common class mIoU score of about $+8.65\%$ compared to the HRDA~\cite{hoyer2022hrda}. This showed that our proposed method not only improved the performance of the private class but also contributed to a slight improvement in the common classes. 
This is because DECON loss encouraged features of the private class near the boundary to converge while distancing themselves from features of the common class. This reduced confusion between the common and private classes, improving predictions of the common class. Moreover, since OpenReMix was designed to learn size-invariant features regardless of the common and private classes, it enhanced the accuracy of predicting the shape of both common and private classes.
We also compared with BUDA~\cite{bucher2021handling}. Since BUDA had access to pre-defined private category definitions and direct comparison was not practical, we offered a comparative analysis in supplementary material.


\subsection{Qualitative Evaluation}
To validate the performance of our method, we conducted additional qualitative evaluations to assess segmentation performance against baselines. We compared our method with MIC, confidence-based MIC (Config. A), and the head-expansion approach (Config. B) in the GTA $\rightarrow$ Cityscapes (see Figure~\ref{fig:qu1}). Furthermore, we compared our method with the head-expansion approach (Config. B), the incorporation of a new DECON loss (Config. C), and the utilization of the new OpenReMix (Config. D) in the SYNTHIA $\rightarrow$ Cityscapes (see Figure~\ref{fig:qu2}).
In Figure~\ref{fig:qu1}, we observed that the UDA method MIC, which was designed for UDA without considering unknown classes, struggled to detect the private classes in OSDA-SS. Even baselines like confidence-based MIC (Config. A) and head-expansion (Config. B) faced challenges in identifying private classes. Although head-expansion showed promise, it still had limitations in classifying specific pixels in private classes. In contrast, our method excelled, particularly in discerning object size.
In Figure~\ref{fig:qu2}, our proposed DECON loss and OpenReMix yielded outstanding performance.



\begin{table}[t]
    \setlength{\tabcolsep}{2.2pt}
    \renewcommand{\arraystretch}{1.6}
    \resizebox{\columnwidth}{!}{
    \begin{NiceTabular}{cccc|ccc}
        \specialrule{2.pt}{1pt}{1pt}
        \multicolumn{4}{c|}{Method}                                     & \multicolumn{3}{c}{GTA5 $\rightarrow$  Cityscapes} \\ \hline
        \multicolumn{1}{c|}{Config.} & \# Head      & DECON & OpenReMix & Common     & Private     & H-Score    \\ \hline
        \multicolumn{1}{c|}{A}       & \textit{C}   &       &           & 58.17      & 11.87       & 19.71      \\
        \multicolumn{1}{c|}{B}       & \textit{C}+1 &       &           & 70.37      & 31.78       & 43.79      \\ \dashedline
        \multicolumn{1}{c|}{C}       & \textit{C}+1 &  \multicolumn{1}{c}{\checkmark}     &           & 71.16      & 48.34       & 57.57      \\
        \multicolumn{1}{c|}{D}       & \textit{C}+1 &       &    \multicolumn{1}{c}{\checkmark}       & 71.52      & 49.26       & 58.34      \\
        \rowcolor[gray]{.9} \multicolumn{1}{c|}{\textbf{Ours}}    & \textit{C}+1 &   \multicolumn{1}{c}{\checkmark}    &    \multicolumn{1}{c}{\checkmark}       & \textbf{72.47}      & \textbf{55.42}       & \textbf{62.81}  \\
        \specialrule{2.pt}{0.1pt}{1pt}
    \end{NiceTabular}%
    }
    \vspace{-2mm}
    \caption{Ablation study of the components in our BUS framework. Configuration A, B, C, and D represent confidence-based MIC, head-expansion, DECON loss, and OpenReMix, respectively.
    \vspace{-2.5mm}}
    \label{tab:ablation1}
    \vspace{-4mm}
\end{table}

\subsection{Ablation Study}

\paragraph{Ablation Study.}
We conducted an ablation study for the proposed components of the BUS framework on GTA5 $\rightarrow$ Cityscapes. In Table~\ref{tab:ablation1}, row A and B represented the confidence-threshold and head-expansion baselines, respectively. The confidence-threshold baseline (Config. A) recorded inferior performance compared to the head-expansion baseline (Config. B). It revealed that leveraging the expanded head was effective in detecting unknown classes,
achieving H-Scores from $19.71\%$ to $43.79\%$. When we combined DECON loss with the head-expansion, we achieved a +$13.78\%$ improvement in the H-Score (see row C). We also confirmed the effectiveness of our proposed OpenReMix. We gained a +$14.55\%$ improvement in the H-Score (see row D).
Lastly, using both DECON and OpenReMix on head-expansion significantly improved H-Score of +$19.02\%$. Figure~\ref{fig:qu2} showed a clear improvement in predicting the unknown compared to the MIC with head-expansion approach (Config. B), and we observed synergy in overcoming individual drawbacks when compared to DECON loss (Config. C) and OpenReMix (Config. D).




\begin{table}[t]
\vspace{0.3mm}
\resizebox{\columnwidth}{!}{%
\setlength{\tabcolsep}{10.pt}
\renewcommand{\arraystretch}{1.2}
\begin{NiceTabular}{c|c|c|c}
\specialrule{1.5pt}{0.1pt}{1pt}
\multicolumn{4}{c}{GTA5 $\rightarrow$ Cityscapes}           \\ \hline
\multicolumn{1}{c}{\# of Unknown} \vline & Config. A & Config. B & Ours  \\ \hline
\multicolumn{1}{c}{6}       \vline      & 19.71          & 43.79              & \multicolumn{1}{c}{\textbf{62.81}} \\
\multicolumn{1}{c}{4}       \vline      & 11.73          & 41.54              & \multicolumn{1}{c}{\textbf{54.72}} \\
\multicolumn{1}{c}{2}       \vline      & 9.43           & 41.51              & \multicolumn{1}{c}{\textbf{56.82}} \\

\specialrule{1.5pt}{0.2pt}{1pt}
\end{NiceTabular}%
}
\vspace{-3mm}
\caption{\label{tab:numof_unknown} The comparison of the number of unknown classes. Config. A denotes the confidence-based MIC and Config. B denotes the head-expansion baseline.
\vspace{-2mm}}
\vspace{-5mm}
\end{table}
\paragraph{Unknown Proportion.}
We conducted the experiments under a various number of unknown classes. When the number of unknown classes was 6, 4, and 2, we compared our method with two MIC-based baselines. For the case of 4 unknown classes, we selected (``pole", ``traffic sign", ``person", ``rider"), and for the case of 2 unknown classes, we chose (``person", ``rider"). Table~\ref{tab:numof_unknown} showed that our proposed method consistently outperformed the baselines, regardless of the change in the number of unknown classes.
\section{Conclusion}
   \label{sec:conclusions}
To tackle this challenging OSDA-SS task, we proposed a novel method named BUS. Our approach includes DECON loss, a new dilation-erosion-based contrastive loss designed to rectify less confident and erroneous predictions near class boundaries. In addition, we proposed OpenReMix guiding the model to acquire size-invariant features and efficiently train the expanded head by mixing unknown objects from the target into the source. 
Through extensive experiments, we demonstrated the efficacy of our proposed method on public benchmark datasets, surpassing existing approaches by a significant margin. We anticipate that our work will be widely applied in research or the industry field, providing a strong baseline to detect unexpected and unseen objects in mission-critical scenarios.
As a limitation, our method is primarily based on pseudo-labeling. Therefore, if the model is poorly calibrated, it might not assign pixels belonging to the private classes as unknown. In this case, BUS might show a performance drop.
\section*{Acknowledgment}
   \label{sec:acknowledgment}
This work was supported by MSIT (Ministry of Science and ICT), Korea, under the ITRC (Information Technology Research Center) support program (IITP-2024-RS-2023-00258649) supervised by the IITP (Institute for Information \& Communications Technology Planning \& Evaluation), and in part by the IITP grant funded by the Korea Government (MSIT) (Artificial Intelligence Innovation Hub) under Grant 2021-0-02068, and by the IITP grant funded by the Korea government (MSIT) (No.RS-2022-00155911, Artificial Intelligence Convergence Innovation Human Resources Development (Kyung Hee University)).


\clearpage
\setcounter{page}{1}
\maketitlesupplementary

%
%
%





\appendix
\setcounter{section}{0}
\setcounter{figure}{0}
\setcounter{table}{0}

\begin{appendix}
\section{Implementation Details}
\begin{figure}[t] 

\centering

\includegraphics[width=\columnwidth]{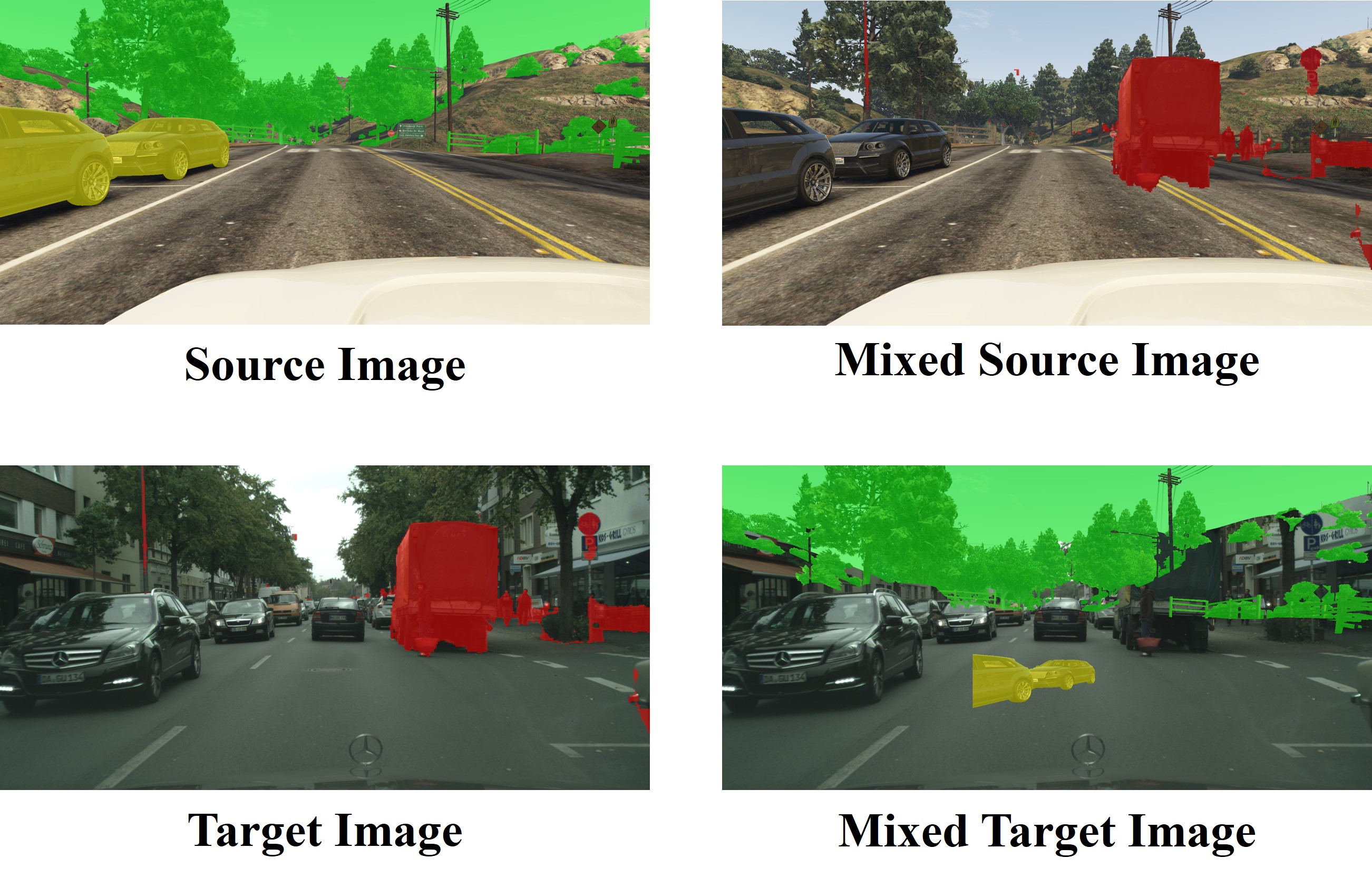}
\vspace{-8mm}
\caption{Example of OpenReMix. The source image is mixed with private classes from the target image (red mask). The target image is mixed by Classmix~\cite{olsson2021classmix} (green mask) and is additionally mixed with an additional resized thing class from the source image (yellow mask).
}

\label{fig:OpenReMix}

\end{figure}

In this section, we provide further implementation details of the proposed method. 
For DECON loss, we crop the target private map to a size of 64$\times$64. Then, we apply the dilation and erosion function. The results of the dilation and erosion functions vary depending on the kernel size and iterations. In this study, we utilize 3$\times$3 kernel size and 1 iteration. 
For OpenReMix, we select one thing class from the source image, resize it, and paste it to the random location of the target image. Here, we resize the selected class by a ratio of 0.5 with bilinear downsampling. We represent an example of OpenReMix in Figure~\ref{fig:OpenReMix}. The resized thing class is marked with a yellow mask. In the attaching private process, the parts predicted as unknown from the target image are attached to the source image. That parts are indicated with a red mask.
Additionally, we utilize MobileSAM~\cite{zhang2023faster} as a refinement network, which is a lightweight version of the Segment Anything Model (SAM)~\cite{kirillov2023segment} for image segmentation. MobileSAM is a highly generalized image segmentation model that can provide reasonable masks for objects in an image even in zero-shot scenarios, but it cannot provide labels. Leveraging these label-less but precise masks, we refine the pseudo-labels. For each generated mask, the pixel count for each class is calculated, and the region of the mask is replaced entirely with the most frequent class. We apply the last 3k iterations every 10k iterations, resulting in a total of 12k iterations out of 40k iterations. And, we also apply the attaching private process in OpenReMix only when pseudo-label refinement is applied.


\section{Hyperparameter Sensitivity}

\subsection{Crop Size in DECON Loss}
We randomly crop the target private mask and apply the dilation and erosion operation for DECON loss. Table~\ref{tab:crop_size} shows the experimental results on the effect of the crop size. In terms of the H-Score, we confirm the robust performance across different crop sizes. And it shows the best performance when cropped to a size of 64$\times$64. Additionally, we observe that the performance significantly decrease in the case of 128$\times$128. This is because, when too much target private information is included in the mask, the anchor cannot reflect the specific characteristics of a particular target private class.

\subsection{Kernel Size in DECON Loss}
We examine the influence of different kernel sizes when applying dilation and erosion functions for DECON loss. In Table ~\ref{tab:kernel_size}, we increase the size from $3 \times 3$  to $7 \times 7$. We confirm that as the kernel size increases, the performance decreases for both scenarios. As the kernel size increases, it considers features further away from the boundary. Therefore, it hinders the model from focusing on the boundary regions where it is difficult to distinguish between known and unknown classes.

\begin{table}[]
\centering
\resizebox{\columnwidth}{!}{%
\setlength{\tabcolsep}{0.5pt}
\renewcommand{\arraystretch}{1.0}
{\scriptsize
\begin{NiceTabular}{c|c|c}
\specialrule{1.5pt}{0.1pt}{0pt}
\multicolumn{1}{c|}{Crop Size}                    & GTA5 $\rightarrow$ Cityscapes & SYNTHIA $\rightarrow$  Cityscapes \\ \hline

$32 \times 32$         &  61.22            &39.39                 \\
\textbf{64 $\times$ 64}   &  \textbf{62.81}             &   \textbf{44.01}              \\
$128 \times 128$ & 61.30              &  37.62                \\ \hline
\specialrule{1.0pt}{0.2pt}{1pt}
\end{NiceTabular}%
}
}
\vspace{-2mm}
\caption{\label{tab:crop_size} Sensitivity of crop size in DECON loss.}
\end{table}

\begin{table}[t]
\centering
\resizebox{\columnwidth}{!}{%
\setlength{\tabcolsep}{0.5pt}
\renewcommand{\arraystretch}{1.0}
{\scriptsize
\begin{NiceTabular}{c|c|c}
\specialrule{1.5pt}{0.1pt}{0pt}
\multicolumn{1}{c|}{Kernel Size}                    & GTA5 $\rightarrow$ Cityscapes & SYNTHIA $\rightarrow$  Cityscapes \\ \hline

\textbf{3 $\times$ 3}         &  \textbf{62.81}             & \textbf{44.01}                \\
$5 \times 5$   &  60.40             &   35.26               \\
$7 \times 7$ & 58.57              &  33.37                \\ \hline
\specialrule{1.0pt}{0.2pt}{1pt}
\end{NiceTabular}%
}
}
\vspace{-2mm}
\caption{\label{tab:kernel_size} Sensitivity of kernel size in DECON loss.}
\end{table}

\begin{table}[t]
\centering
\resizebox{\columnwidth}{!}{%
\setlength{\tabcolsep}{0pt}
\renewcommand{\arraystretch}{1.0}
{\scriptsize
\begin{NiceTabular}{l|c|c}
\specialrule{1.5pt}{0.1pt}{0pt}
\multicolumn{1}{c|}{$s_r$}                    & GTA5 $\rightarrow$ Cityscapes & SYNTHIA $\rightarrow$  Cityscapes \\ \hline
\textbf{[}$\mathbf{0.5,0.5}$\textbf{]}           & \textbf{62.81}              & \textbf{44.01} \\ \dashedline

$[0,2]$ (All)          & 59.73              & 36.77                 \\
$[1,2]$ (Upscale)   & 54.36              &   37.73               \\
$[0,1                                              
                                                   
 ]$ (Downscale) & 59.84              &  38.90                \\ \hline
\specialrule{1.0pt}{0.2pt}{1pt}
\end{NiceTabular}%
}
}
\vspace{-2mm}
\caption{\label{tab:random_resize_scale} Sensitivity of resizing scale in OpenReMix.}
\vspace{-2mm}
\end{table}

\begin{table*}[t]
\centering
\resizebox{\textwidth}{!}{%
\setlength{\tabcolsep}{2.pt}
\renewcommand{\arraystretch}{1.2}

\begin{NiceTabular}{l|ccccccccccccc|ccc}
\specialrule{1.5pt}{1pt}{0pt}
\multicolumn{1}{l|}{Method}                                & Road  & S.walk & Build. & Wall  & Fence & Light & Veget. & Terrain & Sky   & Car   & Bus   & M.bike & \multicolumn{1}{c|}{Bike}  & Common & Private & H-Score \\ \hline
\multicolumn{17}{c}{\textbf{GTA5 $\rightarrow$ Cityscapes}}                                                                                                                                                                 \\ \hline
\multicolumn{1}{l}{DAF~\cite{hoyer2022daformer}}         & 95.80 & 65.37  & 87.12  & 54.08 & 45.81 & 51.78 & 89.20  & 42.93   & 91.03 & 89.19 & 37.93 & 50.54  & \multicolumn{1}{c}{48.49} & 66.09  & 29.23   & 40.53   \\
\rowcolor[gray]{.9}\multicolumn{1}{l}{DAF + BUS}   & 91.90 & 41.06  & 88.04  & 48.65 & 48.74 & 48.94 & 89.59  & 44.37   & 91.61 & 89.99 & 46.09  & 48.49   & \multicolumn{1}{c}{62.47}  &  64.61 & 39.23    & 48.82     \\
\multicolumn{1}{l}{HRDA~\cite{hoyer2022hrda}}             & 95.31 & 37.70  & 89.26  & 57.41 & 37.00 & 61.16 & 90.96  & 46.86   & 94.39 & 93.39 & 62.45 & 58.13  & \multicolumn{1}{c}{65.71} &  68.44 & 31.02   & 42.70   \\
\rowcolor[gray]{.9}\multicolumn{1}{l}{HRDA + BUS}       & 88.07 & 39.59  & 88.57  & 55.12 & 48.29 & 56.24 & 90.02  & 46.30   & 91.76 & 92.03 & 46.96 & 57.10  & \multicolumn{1}{c}{66.02} &  66.62 & 42.50   & 51.89   \\
\multicolumn{1}{l}{MIC~\cite{hoyer2023mic}}              & 97.14 & 79.45   & 88.78  & 55.6  & 53.92  & 26.11 & 89.94  & 50.98   & 93.54 & 92.46 & 69.09 & 54.53  & \multicolumn{1}{c}{63.43}  & 70.38  & 31.78    & 43.79   \\
\rowcolor[gray]{.9}\multicolumn{1}{l}{MIC + BUS (Ours)} & 95.06 & 66.65  & 90.53  & 55.37 & 55.38 & 57.20 & 91.12  & 49.69   & 92.96 & 93.50 & 68.81 & 58.73  & \multicolumn{1}{c}{67.04} &  72.47  &  55.42   &  62.81   \\
\specialrule{1.5pt}{1pt}{1pt}
\end{NiceTabular}%
}
\vspace{-2mm}
\caption{\label{tab:others} Comparison with some self-training-based UDA methods. White row denotes the head-expansion baseline and gray row means our proposed BUS.}
\end{table*}

\subsection{Resizing Scale in OpenReMix}
We provide the results on various resizing factors for OpenReMix in Table~\ref{tab:random_resize_scale}. For each iteration, we randomly select the scale factor from a uniform distribution within a specified range. From this result, we confirm that the proposed OpenReMix is robust to scale factors, and a simply fixed scale factor of 0.5 is enough to learn size-invariant features for our model.

\subsection{Threshold in Pseudo-Label Generation}
We study the influence of different thresholds $\tau_p$ for assignment of unknown classes during pseudo label generation. Table ~\ref{tab:threshold} shows the results under the various values of $\tau_p$ in GTA5 $\rightarrow$ Cityscapes scenario. We observe that for any value other than $\tau_p=0.5$, the performance degrades significantly. Therefore, our method is sensitive to $\tau_p$, so selecting an appropriate threshold is  important.

\begin{table}[t]
\centering
\resizebox{\columnwidth}{!}{%
\setlength{\tabcolsep}{8.pt}
\begin{tabular}{c|c|c|c|c|c}
\specialrule{1.5pt}{0.1pt}{0pt}
$\tau_p$     & 0.3   & 0.4   & \textbf{0.5}   & 0.6   & 0.7   \\ \hline
H-Score & 17.91 & 32.21 & \textbf{62.81} &26.74  & 23.36 \\ \hline
\specialrule{1.0pt}{0.1pt}{1pt}
\end{tabular}%
}
\vspace{-2mm}
\caption{\label{tab:threshold} Sensitivity of threshold $\tau_p$ in GTA5 $\rightarrow$ Cityscapes scenario.}
\end{table}



\begin{table}[]
\centering
\setlength{\tabcolsep}{20.5pt}
\resizebox{\columnwidth}{!}{%
\begin{tabular}{c|cc>{\columncolor[gray]{0.9}}c}
\specialrule{1.0pt}{0.1pt}{0pt}
\hline
Method       & Common        & Private       & H-Score       \\ \hline
BUDA         & 37.3          & 18.5          & 24.7          \\
MIC          & 54.3          & 24.1          & 34.4          \\
\textbf{BUS} & \textbf{55.6} & \textbf{39.7} & \textbf{46.3} \\ \hline
\specialrule{1.0pt}{0.2pt}{1pt}
\end{tabular}%
}
\vspace{-2mm}
\caption{\label{tab:BUDA_experiment} Comparison with BUDA in Cityscapes $\rightarrow$ IDD scenario.}
\vspace{-4mm}
\end{table}

\section{Comparison with Other Baselines}
Our proposed methods can be applied to existing self-training-based UDA methods. Therefore, we present the results applying the head expansion baseline to existing UDA methods, as well as the results incorporating the two components we propose, which are DECON loss and OpenReMix. In Table~\ref{tab:others}, we confirm  the increase of H-Score for DAF~\cite{hoyer2022daformer} by $+8.29\%$ and for HRDA~\cite{hoyer2022hrda} by $+9.19\%$. Particularly, in the case of MIC~\cite{hoyer2023mic}, there was a substantial increase of $+19.02\%$. We confirm that the better the performance of UDA, the better the performance when applying our proposed methods. This is because DECON loss and attaching private process are based on the quality of pseudo labels. Therefore, the models that generate more accurate pseudo-labels have an advantage.

We also compare with the most similar work BUDA~\cite{bucher2021handling} to our method.
In BUDA, models have access to \emph{private category definitions}, a crucial assumption not shared by OSDA-SS. In our OSDA-SS setting, there is no provision for such private category definitions. 
In OSDA-SS, one should devise a method that \emph{rejects} novel classes without needing to know any information about their definition. In BUDA, one should devise a method that \emph{predicts} novel classes explicitly \emph{at the expense of} predefined class definitions.
Given the fundamental differences between OSDA-SS and BUDA, direct comparison is not practical. Nonetheless, we offer a comparative analysis in Table~\ref{tab:BUDA_experiment}. 
To demonstrate the applicability of our proposed methods to various datasets, we conduct experiments on a new dataset called IDD (India Driving Dataset).
Please note that BUDA has the \emph{privilege to access novel class definitions} while BUS do not.

\begin{table}[t]
\vspace{0.3mm}
\resizebox{\columnwidth}{!}{%
\setlength{\tabcolsep}{15.pt}
\renewcommand{\arraystretch}{1.2}
\begin{NiceTabular}{c|c|c|c}
\specialrule{1.5pt}{0.1pt}{0pt}
\multicolumn{4}{c}{\textbf{GTA5} $\rightarrow$ \textbf{CityScapes}}           \\ \hline
\multicolumn{1}{c}{\# of Unknown} \vline & Config. A & Config. B & \textbf{Ours}  \\ \hline
\multicolumn{1}{c}{6}       \vline      & 19.71          & 43.79              & \multicolumn{1}{c}{62.81} \\
\multicolumn{1}{c}{8}       \vline      &   20.06        &        52.76       & \multicolumn{1}{c}{62.01} \\
\multicolumn{1}{c}{10}       \vline      &   18.89         &        48.88       & \multicolumn{1}{c}{55.56} \\

\specialrule{1.5pt}{0.2pt}{1pt}
\end{NiceTabular}%
}
\vspace{-2mm}
\caption{\label{tab:numof_unknown_up} Experiments of different private classes. Config. A denotes confidence-based MIC and config. B denotes MIC with head-expansion.}
\end{table}

\begin{table}[t]
\centering
\setlength{\tabcolsep}{10.5pt}
\resizebox{\columnwidth}{!}{%
\begin{tabular}{c|cc>{\columncolor[gray]{0.9}}c}
\specialrule{1.0pt}{0.1pt}{0pt}
\hline 
Method & Common              & Private              & H-Score             \\ \hline
MIC                     & 60.35 \(\pm\) 6.55          & 61.38 \(\pm\) 10.61          & 59.66 \(\pm\) 3.33         \\
\textbf{BUS}           & \textbf{64.16 \(\pm\) 7.07} & \textbf{66.22 \(\pm\) 11.89} & \textbf{64.33 \(\pm\) 3.45} \\ \hline
\specialrule{1.0pt}{0.2pt}{1pt}
\end{tabular}%
}
\vspace{-2mm}
\caption{\label{tab:random_private} Experiments on randomly selected private categories. We conducted three experiments and presented the average deviation.}
\end{table}

\begin{table}[t]
\centering
\setlength{\tabcolsep}{18.5pt}
\resizebox{\columnwidth}{!}{%
\begin{tabular}{c|c|c|c|c}
\specialrule{1.5pt}{0.1pt}{0pt}
$N$     & \textbf{1}   & 3     & 6   & 10   \\ \hline
H-Score & \textbf{62.81} & 46.02 &  38.01 & 30.56 \\ \hline
\specialrule{1.0pt}{0.1pt}{1pt}
\end{tabular}%
}
\vspace{-2mm}
\caption{\label{tab:N_C_Head} Influence of the number of expanded head in GTA5 $\rightarrow$ Cityscapes.}
\end{table}

\section{More Experiments about Private Classes}
In the main paper, we experimented with a total of six private classes in the GTA $\rightarrow$ Cityscapes scenario and included results for scenarios where the number of private classes decreases. In Table~\ref{tab:numof_unknown_up}, we further present the comparison results when the number of private classes increase. For 8 private classes, we include (``M.bike", ``Bike"), and for 10 unknown classes, we additionally add (``Light", ``Bus"). Despite an increase of the number of private classes, our method still outperform the other baselines. 

As we mentioned in the scenario construction section in main paper, we selected private classes from the thing categories. 
While it is rare for stuff classes to emerge in the real world, we conduct experiments on cases where stuff classes are also treated as private classes.
We randomly select 6 privates out of 19 classes regardless of thing and stuff categories in GTA5 $\rightarrow$ Cityscapes scenario. Table~\ref{tab:random_private} demonstrates that BUS still outperforms the previous baseline in various settings with a significant margin.

\section{What if using $(C+N)$ heads?}
In our OSDA-SS task, the number of private classes $N$ is unknown since target private labels are absent. In that sense, setting $N=1$ for the private class is a reasonable option. Despite this, we experiment using $(C+N)$ heads with random pseudo-labeling. Understandably, Table~\ref{tab:N_C_Head} demonstrates that our BUS shows the best performance when $N$ is set to 1.

\end{appendix}
{
    \small
    \bibliography{main}
}

\end{document}